\documentclass[runningheads]{llncs}
\usepackage{graphicx}
\usepackage{amsmath,amssymb} 
\usepackage{color}
\usepackage{array,multirow}
\usepackage{enumitem}
\usepackage[final]{pdfpages}
\usepackage{bm}
\usepackage[width=122mm,left=12mm,paperwidth=146mm,height=193mm,top=12mm,paperheight=217mm]{geometry}
\usepackage{amsmath}
\usepackage{wrapfig}
\usepackage{lipsum}  
\DeclareMathOperator*{\argmax}{argmax} 

\begin{document}
\pagestyle{headings}
\mainmatter
\def\ECCV18SubNumber{960}  

\title{Real-time Burst Photo Selection Using a Light-Head Adversarial Network } 



\author{Baoyuan Wang,
        Noranart Vesdapunt,
        Utkarsh Sinha, Lei Zhang}
\institute{Microsoft Research}

\maketitle

\begin{abstract}
We present an automatic moment capture system that runs in real-time on mobile cameras. The system is designed to run in the viewfinder mode and capture a burst sequence of frames before and after the shutter is pressed. For each frame, the system predicts in real-time a ``goodness'' score, based on which the best moment in the burst can be selected immediately after the shutter is released, without any user interference. To solve the problem, we develop a highly efficient deep neural network ranking model, which implicitly learns a ``latent relative attribute" space to capture subtle visual differences within a sequence of burst images. Then the overall goodness is computed as a linear aggregation of the goodnesses of all the latent attributes. The latent relative attributes and the aggregation function can be seamlessly integrated in one fully convolutional network and trained in an end-to-end fashion. To obtain a compact model which can run on mobile devices in real-time, we have explored and evaluated a wide range of network design choices, taking into account the constraints of model size, computational cost, and accuracy. Extensive studies show that the best frame predicted by our model hit users' top-1 (out of 11 on average) choice for $64.1\%$ cases and top-3 choices for $86.2\%$ cases. Moreover, the model(only 0.47M Bytes) can run in real time on mobile devices, e.g. only 13ms on iPhone 7 for one frame prediction.
\end{abstract}

\begin{figure*}[t]
\centering
\includegraphics[width=\linewidth]{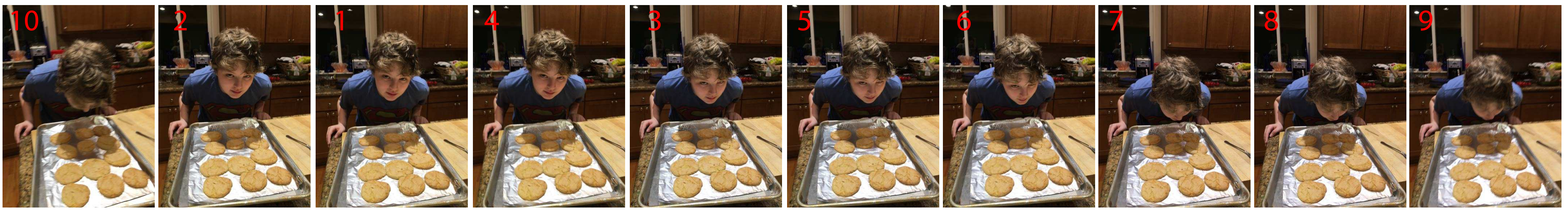}
\caption{An example of burst frames and their ranking predicted by our system, which jointly considers the relative sharpness, facial expression as well as the head pose. Note that, even though the $8$-th frame(from left to right) is very sharp, its rank is still very low because of the head pose. Best viewed in a zoomed-in  electronic version.}
\label{fig:teaser}
\end{figure*}

\section{Introduction}
This paper addresses the problem of how to take pictures of the best moment using mobile phones. With the recent advances in hardware, such as Dual-Lens camera on iPhone 7 Plus, the quality of the pictures taken on mobile phones has been dramatically improved. However, capturing a great ``moment" is still quite challenging for common users, because anticipating the subject movements patiently while keeping the scene framed in viewfinder requires lots of practices and professional training. For example, taking spontaneous shots for children could be extremely hard as they may easily run out of the frame by the time you press the shutter. As a result, one may not only miss the desired moment, but also get a blurry photo due to the camera or subject motion. Taking another common example in portrait photography, keeping a perfect facial expression for long time without blinking eyes is nearly impossible. Therefore,  it is likely that one has to replicate his pose and expression multiple times in order to capture a perfect shot, or one can use the burst mode to shot dozens of photos and then manually select the best one to keep and discard the rest. Although this method works for some people, it is less efficient due to the fact of wasting storage space and intensive manual selection.


In this paper, we introduce a real-time system that automates the best frame (great moment) selection process ``during" the capture stage without any post-capture manual operations. Specifically, we propose to buffer a few frames before and another few frames after the shutter press, we then apply an efficient photo ranking model to surface the best moment and automatically remove the rest of them to save storage space. We argue that having a real-time capture system would dramatically lower the bar of high quality moment capture for memory keeping or social sharing.

To our best knowledge, there is no prior work in academia that directly targets at building automatic moment capture system during the \textit{capture} stage, not to say on mobile phones. This is mainly due to the following challenges. First, such a system needs to run during the capture stage in the viewfinder, the ranking model has to be compact enough to be deployed on mobile phones and fast enough to run in real-time. Second, learning such an efficient and robust ranking model is challenging because the visual differences within a sequence of burst images are usually very subtle, yet the criteria for relative ranking could range from low-level image quality assessment, such as blur and exposure, to high-level image aesthetics, such as the attractiveness of facial expression or body pose, requiring a holistic way of learning all such representations in one unified model. Last but not least, due to the uniqueness of this problem, there is no available burst image sequences to serve as our training data, and it is also unclear how to collect such supervision signals in an effective way. For the same reasons, we cannot leverage related works developed for automatic photo selection from personal photo albums, because their photo selection criteria primarily focus on \textit{absolute} attributes such as low-level image quality \cite{wang2004image}, memorability \cite{IsolaParikhTorralbaOliva2011}, popularity \cite{Khosla:2014:MIP}, interestingness \cite{Fu_interestingnessprediction}, and aesthetics \cite{Aesthetics2011}. In contrast, we are more interested in learning \textit{relative} attributes that can rank a sequence of burst images with subtle differences. 

To address these challenges, we first created a novel burst dataset by manually capturing 15k bursts covering a rich set of common categories including selfies, portrait, landscaping, pets, action shots and so on. We sample image pairs from each burst and then conducted crowd-sourcing through Amazon Mechanical Turk (AMT) to get their overall relative goodness label (i.e., which looks better?) for each image pair. We consolidate the label information by a simple average voting. Second, considering a pair of images sampled from a burst, the visual content is largely overlapped, indicating the high-level features of a convolution network pre-trained for image classification may not be suitable for relative ranking, because classification network generally tries to achieve certain translation and rotation invariance and be robust to certain degree of image quality variations for the same object. However, those variances are the key information used for photo ranking. Therefore, in order to leverage the transfer learning from an existing classification net, one can only borrow the weights of the backbone net\footnote{Those weights in backbone will be fine-tuned when training the ranking net} and must re-design a new head to tailer for our photo ranking problem.
In addition to this, we observed that the relative ranking between a pair of images is determined by a few relative attributes such as sharpness, eye close or open, attractiveness of body pose or overall composition. And the overall ranker should be an aggregation of all such  relative attributes. To enforce this observation, also inspired by recent advances in Generative Adversarial Networks(GANs)\cite{GaN0,GAN1,CGAN14}, we introduce another generator (denoted as ``G'') that can enhance the representation of the latent attributes so as to augment more training pairs in the feature space for improving the ranking model. Although we do not have the attribute level label information during the training, we expect the ranking network with a novel head can learn latent attribute values implicitly, so that it can minimize the ranking loss more easily. 

Motivated by the above facts and observations, we explored various choices for the backbone network and head (the final multi-layer module for ranking) design, and proposed a compact fully convolution network that can achieve good balance among model size, runtime speed, and accuracy. To sum up, we made the following contributions:
\begin{itemize}[font=\bfseries]
\item{We propose an automatic burst moment capture system running in real-time on mobile devices. The system can save significant storage space and manual operations of photo selection or editing for end users.}
\item{We explored various network backbone and head design choices, and developed a light-head network to learn the ranking function. We further applied the idea of Generative Adversarial Networks(GANs) into our framework to perform feature space augmentation, which consistently improves the performance for different configurations. }
\item{We deployed and evaluated our system on several mobile phones. Extensive ablation studies show that our model can hit $64.1\%$ user's top-1 accuracy(out of 11 on average). Moreover, the model(0.47M Bytes) can run in real time on mobile devices, e.g. only 13ms on iPhone 7 for one frame prediction.}
\end{itemize}

\section{Related Works}
\subsubsection{Automatic Photo Triage}
Automatic photo selection from personal photo collections has been actively studied for years \cite{Chu:2008:ASR,Ceroni:2015:KKE,Walber:2014,Sinha:2011:SPP}. The selection criteria, however, are primarily focused on low-level image quality, representativeness, diversity, as well as coverage. Recently, there has been an increasing interest in understanding and learning various high-level image attributes, including memorability \cite{IsolaParikhTorralbaOliva2011,Isola2011,GygliICCV13,ICCV15_Khosla,Dubey_2015_ICCV},  popularity \cite{Khosla:2014:MIP}, interestingness \cite{Fu_interestingnessprediction,GygliICCV13,Aesthetics2011,Dhar:2011:HLD}, aesthetics \cite{Aesthetics2011,Lu:2014:RRP,Datta:2006:SAP,Dhar:2011:HLD}, importance \cite{Importance2012} and specificity \cite{Jas_2015_CVPR}. So technically, photo triage could be alternatively solved by assessing each of those image attributes. Although these prior works are relevant, our work is distinct in a number of ways: (1) we are interested in learning the ranking function that only runs ``locally" within the burst rather than globally across all bursts. We do not expect the ranker to perform well between different bursts, because images coming from different bursts may not even be comparable; (2) we are interested in learning the ``relative" attribute values. For example, both image A and B are blurry, but A is still relatively less blurry compared with B. However, all these prior works target for learning the attributes in an absolute manner; (3) Our ranker learns all the latent attributes (sharpness, smile, eye open/close etc) holistically in a weakly supervised manner while prior works all deal with each individual attribute with full supervision. There are a few interesting works along the line of relative attribute learning, such as \cite{conf/iccv/ParikhG11,Xiao-iccv2015,DBLP:journals/corr/SouriNM15,Relative_Parts}. Yet they are not designed to rank photos for moment capture and require full supervision to train each individual attribute independently. 

Technically, the automatic photo triage system proposed by Chang etc ~\cite{Chang:2016:ATF} might be the only work close to ours. However, it does not support burst photos for moment selection, as the differences are too subtle to treat each burst session as a ``photo series'' as defined in their setting. Moreover, we argue that their proposed network design is less efficient for real-time burst moment capture, as the ranker always need to feed the feature differences of an image pair to return the winner, and the winner is then recursively paired with the next frame until it loops over the whole burst. Clearly, this process can't be easily ran in parallel. Especially, the complexity of getting the full rank is $O(n^2)$ ($n$ is the number of frames within the burst). In contrast, our ranker requires only one frame as input and directly predict its goodness score for ranking.

\subsubsection{Learning to Rank} In the domain of image retrieval \cite{Image_ReRank11,ImageRank,ImageRetrieval07}, the ranking functions generally associate with a query image and the goal is to rank the relevance of the resulting images with respect to the query. Whereas in our setting, as there is no query or reference image associated, our ranking function has to learn the degree of image goodness that can be determined by a few latent relative attributes .
\subsubsection{Generative Adversarial Networks(GANs)} GAN \cite{GaN0} and its variants have become hot in research due to their ability of generating new samples from training data. However, most of the prior works\cite{GaN0,GAN1,CGAN14,pix2pix2016} focus on realistic generating tasks, while in our work we use the concept of GAN to perform feature space augmentation to improve a ranking model. We will discard the Generator during the runtime.
\subsubsection{Burst Moment Capture}
To our best knowledge, there is no public prior work that directly targets for building an automatic moment capture system during the ``capture'' stage. Commercial product such as Microsoft Pix claims to have similar feature, it is however unclear how they implemented it technically. Most popular native cameras on smart phones such as iPhone, Google Pixel support burst capture by letting users keep pressing the shutter or holding it down for a while. The users then navigate to the photo album and manually compare all frames to pick the best and discard the rest ones. Although the system usually marks the best frame(s), the results are not always satisfactory.  
\section{Overview}

\subsection{Problem Formulation}\label{Sec:formulation} As mentioned above, we formulate the burst moment capture as a local relative ranking problem. Precisely, given an image burst with $l$ frames, denoted as $S=\{I_0,I_1,...,I_l\}$, our goal is to find a scoring function $f(x)$ that only takes a single frame as input and output its corresponding goodness score for ranking, and the best moment can be simply found by $\argmax_{x \in \mathcal{S}} f(x)$. Note that, as $f(x)$ is trained to rank image pairs only within the same burst, it does not need to worry about the comparison across different burst. Therefore, $f(x)$ is forced to learn the relative attributes that can rank one image higher than the other in an image pair.
\subsection{Data Set}
\subsubsection{Burst Sequences}
As there is no public burst dataset available for our purpose, to train the ranking function $f$, we have to collect our own dataset. One potential idea is to sample continuous frames from existing videos to mimic burst sequences which seems to be fairly easy. However, we have found it nontrivial to collect a large video set with diverse categories to approximate the distribution of generic photographs. Furthermore, the defects existing in burst capture mode may not necessarily be the same as in video mode. Nevertheless, we leave this as the future work for dataset augmentation. 

To start with, we hired ten people to perform the data collection work. They were asked to always use the burst mode to capture each moment they want. Three mobile phones including iPhone SE, Google Pixel and Samsung S8 were used to collect bursts that range from a rich set of categories on purpose. In total, we have collected 14,769 bursts (246,715 images). We then randomly split the whole burst set into three subsets for training(8627), validation(1000) and testing(5142) respectively. On average, there are 11 images from each burst.

\begin{figure*}[t]
\centering
\includegraphics[width=0.95\linewidth]{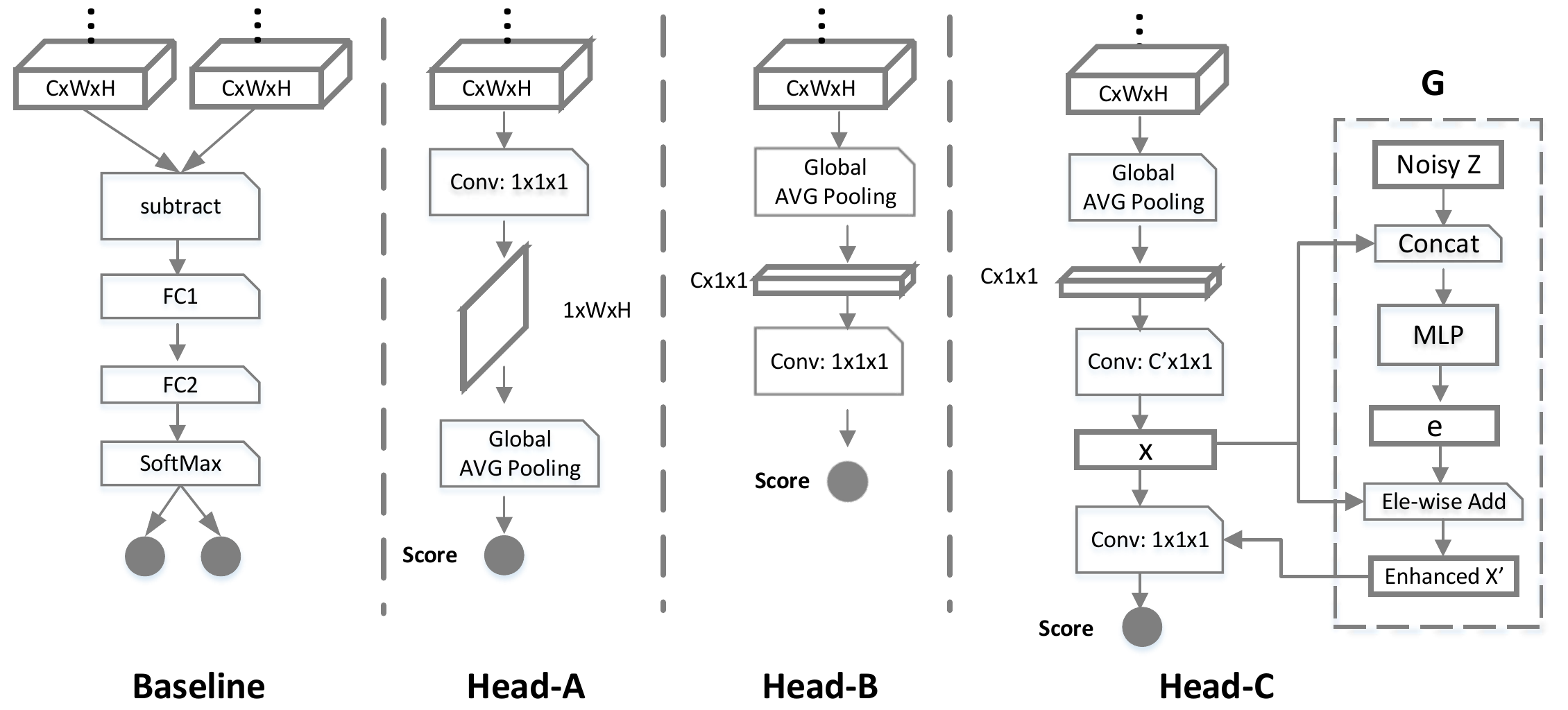}
\caption{Different head design choices that sit on top of any backbone network layer. The dashed box in ``Head-C" indicate the component ``G'' only exists during training. The ``Baseline'' used by \cite{Chang:2016:ATF} always requires two images as input and mainly uses stacked fully connected layers, which is not efficient during runtime}
\label{fig:head_design}
\end{figure*}

\subsubsection{Crowd-Scouring}
We sampled all the image pairs from each burst and conducted an Amazon Mechanical Turk(AMT) study to get pair-wise labels. Instead of annotating the overall image quality in absolute manner, our pair-wise comparison  encourages Turkers to pay attention to the subtle differences. We present one image pair side-by-side each time and ask the Turkers to select the options from (1) ``The Left(Right) image is significantly better than the Right(Left) image" (2) ``The Left(Right) image is marginally better than the Right(Left) image'' (3) ``They are equally good''. Given a sample pair $(A,B)$, we use $ A \succeq B$ to represent that $A$ is equal or better than $B$ (we can switch if B is better), then the label information $Y$ for $ A \succeq B$ can be defined as 
\vspace{-1mm}
\begin{equation} \label{eq1}
Y(A,B) = 
\begin{cases}
0 &\mbox{if A, B are equal}\\
1 &\mbox{if A is marginally better} \\
2 &\mbox{if A is significantly better}
\end{cases}
\end{equation}
Each image pair was judged by 5 different Turkers, the final label was consolidated by averaging $Y$ and then rounded to 0, 1, or 2. Other more advanced consolidation method, such as the modeling probability distribution of the annotations, could be explored in the future work. 

In summary, our dataset consists of 838,038 pairs as well as their corresponding labels. Overall, around $35\%$ of the pairs have equal relative goodness ($\Delta(y)==0$), which is not surprising, because when there is no camera motion or subject motion, such as in still landscape photography, the frames within the same burst tend to be very close to each other.

\subsection{Margin Rescaling Based Ranking Model}
We propose to use a fully convolution network to learn both the discriminative features as well as the scoring function $f$ for ranking. As discussed in Sec. \ref{Sec:formulation}, we need to learn $f(x)$ that can  predict the overall goodness score for an input frame $x$. Although we do not have the direct score label to train a regression function for $f(x)$, we have the pair-wise label that collected though AMT as discussed above.
For a given pair $(A,B)$ from the training set, let us denote $x^i_A$ ($x^i_B$) as the feature of image A (B) in the $i$-th pair, and $\Delta(y_i)=Y(A,B)$ as its pair-wise label ground-truth. Then we need to learn the function $f$ so that for $f(x^i_A)-f(x^i_B) \geq \Delta(y_i)$ if $\Delta(y_i) \neq 0$, otherwise $|f(x^i_A)-f(x^i_B)| \leq \gamma_i$ where $\gamma_i > 0$. Then the loss for the ranking model $\mathcal{R_{\phi}}$ of each pair can be defined as
\begin{equation}\label{equ:loss}
\begin{split}
\mathcal{R_{\phi}}(A \succeq B,\Delta(y)) =\max(0, \Delta(y_i)-(f(x^i_A)-f(x^i_B)))
+|f(x^j_A)-f(x^j_B)|
\end{split}
\end{equation}

%
%

Note that, different from the traditional rank SVM (i.e., \cite{conf/iccv/ParikhG11}), we use the margin rescaling technique (\cite{Joachims:2009:PSO:1592761.1592783,Taskar:2003:MMN:2981345.2981349}) to enforce the ranking function to respect the degree of difference represented as $\Delta(y)$. In our experiments, we observe a slight improvement on our current test set.

\section{Attribute-Aware Head Design}
Intuitively, as we have discussed before, the relative goodness of a pair $(A,B)$ is attributed to a combination of a few relative attributes. For example, from the sharpness perspective, A is slightly better than B, but from the facial expression and image composition perspectives, B is more preferred. Depending on the degree of gap between A and B for each relative attribute, the final relative goodness is determined by a linear combination of multiple relative attributes. Mathematically, $f(x^i_A) - f(x^i_B) = W^T(x^i_A - x^i_B) = \sum w_k(x^{ik}_A-x^{ik}_B)$, where $x^{ik}_A$ is the $k$-th attribute value of image A in the $i$-th pair and $W$ is the combination weighting vector. Even though, in the context of a fully convolution network, the attribute feature vector (i.e., $x_A, x_B$) and its weighting vector $(i.e., W)$ are naturally integrated and can be trained end-to-end, we still expect the network design to be able to respect this simple intuition. We argue that, by adhering to the attribute-aware design guideline, it will be more feasible to learn the intrinsic relative attributes even with a network of small capacity and less computational overhead, which is crucial for real-time moment capture.

One straightforward idea we could try is to customize the head of a typical classification network to output a single score. Following the design principle of several successful fully convolution networks, such as SqueezeNet \cite{squeezenet}, Network in Network \cite{NIN}, one can apply 1x1 point-wise convolution filter first to reduce all the feature maps into one single map with the same spatial size, and then conduct global average pooling to get the final score, as can be illustrated as ``Head-A" in Figure \ref{fig:head_design}. Alternatively, we can flip the operations by first doing global average pooling and then conduct a 1x1 point-wise convolution (same as a fully connected layer) to output the final score\footnote{In our current experiment, except the last scoring layer, there is always a ``ReLu" operation followed after each convolution layer.}. Such design is also popular in classification networks such as Xception \cite{xception} and ResNet \cite{resnet}. We call this design as ``Head-B'' as shown in the middle of Figure \ref{fig:head_design}. Intuitively speaking, the design of ``Head-A" encourages to encode spatial features whereas ``Head-B" encodes the channel features. However, both ``Head-A" and ``Head-B" reduce their spatial feature or channel feature too quickly, which potentially lead to the loss of certain informative features that represent the relative attributes used for subsequent comparison. 

\subsection{Adversarial Ranking Loss Regularization}
To learn a better ranker, the network needs to output a compact set of attribute features before reducing to one final score. Inspired by the prior works on latent topic modeling\cite{TopicModelling} and Generative Adversarial Networks(GANs) \cite{GaN0,GAN1}, we propose a third head design that specially tailored for our relative ranking problem, which is called ``Head-C"  as illustrated in Figure \ref{fig:head_design}. Compared with ``Head-A" and ``Head-B", we first add an extra layer (Conv: $C'\times 1\times 1$) to project the original features into a low-dimensional subspace (denoted as $x$ in Head-C of Figure \ref{fig:head_design}) which can be regarded as a topic space or attribute space. Our hypothesis is that having an intermediate layer before outputting the final score would preserve more informative attribute features. We argue that one can safely assume the final convolution layer (equivalent to a FC layer) in Head-C serves as a linear ranker for all the attribute features from $x$. If we can learn discriminative and robust attribute features, then the ranker becomes easier to train.

Another observation is, for any given image pair $(A,B)$ sampled from a burst, assume $ A \succeq B$ ($f(x_A) \geq f(x_B)$), if the learned intermediate layer in $x$ is indeed attribute aware,  it is likely that by tweaking some attribute values in $x_B$ to get $x_B'$, one could flip the ranking relationship to $ B' \succeq A$ ($f(x_B') \geq f(x_A)$). An intuitive interpretation is that if the reason why $ A \succeq B$ is only caused by the degree of blurriness, then one can just reduce the value of the corresponding blurriness attribute in $x_B$ so that to flip their rank. This inspires us to synthesize more pairs in the attribute space as additional regularization to train the final ranking (scoring) network. 
To do this, we introduce another network which is called ``G" during the training to synthesize a new attribute feature $x'$ for each $x$, as shown in Figure \ref{fig:head_design}, by asking G to output a sparse residual vector $e$ so that $x'=x+e$. Like conditional-GAN \cite{CGAN14}, G takes both $x$ and a randomly generated Gaussian noise vector as input, and then feed into a MLP subnet to output a residual vector $e$. During the training, we add a L1 norm loss constraint to encourage sparsity in $e$ to reduce the risk of over-fitting when training G. So the enhanced $x'_B$ can be regarded as the corresponding attribute feature for a new synthesized image $B'$. If G is well trained, we want to inject a new pair-wise loss for $B'  \succeq B$ when training the main network in Head-C. Compared with traditional Generative Adversarial Network(GAN)\cite{GaN0,GAN1,CGAN14}, we use a ranker instead of a discriminator to drive how we learn the enhanced feature, so training an accurate ranker is our final objective. The purpose of the generator is primarily feature augmentation during the training to help the ranker, whereas in traditional GAN the generator is the major learning objective. Nevertheless, as in traditional GAN, we train both our ranker and G iteratively. Specifically, for each pair $A \succeq B$, when we train G, we want to minimize the loss for enforcing both $f(x_B') \geq f(x_A)$ and the sparsity in $e$; when we train the ranker, in addition to original loss for $f(x_A) \geq f(x_B)$ , we have to add two more losses that enforce both $f(x_B') \geq f(x_B)$ and $f(x_A) \geq f(x_B')$. Mathematically, we will iteratively minimize the following two losses, namely $L(\mathcal{R_{\phi}})$ and $L(\mathcal{G_{\theta}})$ for ranker and G respectively:
\begin{equation}
\min_{\phi} L(\mathcal{R_{\phi}}) = \mathbf{E}_{(A,B)	\sim\mathcal{P}}[\mathcal{R_{\phi}}(A \succeq B,\Delta(y)) +
                \mathcal{R_{\phi}}(B' \succeq B,2)  +\gamma\mathcal{R_{\phi}}(A \succeq B',2)]                
\end{equation}

\begin{equation}\label{equ:g}
\min_{\theta} L(\mathcal{G_{\theta}}) =\mathbf{E}_{(A,B')\sim\mathcal{G_{\theta}}}[\mathcal{R_{\phi}}(B' \succeq A, 2) + \lambda\|e\|_1]
\end{equation}

where $\mathcal{P}$ represents the set of all real pairs in our dataset. $\lambda$ is empirically set to $0.1$, $\gamma$ is initialized to 1 at beginning. We will first train ranker for several epochs without adding the synthetic pairs(i.e.,($A  \succeq B'$) or ($B' \succeq B$)), and train another few epochs for G while fixing the ranker by minimizing Equ. \ref{equ:g}. We then start to train both ranker and G iteratively in a more frequent way, say each 25 mini-batch iterations. In our current experiment, we only take the lower-quality image, i.e, image B, in the pair and feed into G to get an enhanced $B'$. Note that, to enforce the margin, we set $\Delta(y)$ to 2 for all pairs that involved with a synthetic feature. After convergent, we discard G, and continue to fine-tune $\mathcal{R_{\phi}}$ by setting $\gamma$ to zero, assuming $B'$ can be safely ranker higher than $B$.

We argue that our proposed ``Head-C'' with GAN loss is general enough that can sit on top of any backbone, even though we are more interested in studying its effectiveness for small backbone models considering the piratical applications.

\section{Experiments}

\subsection{Evaluation Metric}
\subsubsection{Pair-wise Level} Since we train our system using pair-wise loss, we first measure the pair-wise level accuracy. For each pair $(x^A_i,x^B_i)$ and its ground-truth label $\Delta(y_i)$ in the test set, if $(f(x^A_i)-f(x^B_i))*\Delta(y_i) >0$ we regard the prediction as correct. Note that we have not taken the pairs with equal label into evaluation, as numerically it is infeasible to let $f(x^A_i)=f(x^B_i)$. We care more about the pairs whose $\Delta(y_i)\neq 0$.

\subsubsection{Burst Level} The goal of our system is to predict the best frame from the burst sequence, thus the most interesting metric is to measure the ranking position of our predicted best frame in the list that is sorted based on user's pair-wise label. This motivates us to use Top-K accuracy (with $K=1,2,3 $ in our current experiment). Intuitively, for each burst, according to the labeling, our prediction hits the Top-K accuracy if its rank is less or equal to K in the list sorted based on user label. Specifically, for a burst that contains $l$ frames, the Top-K accuracy is hit, if and only if $\sum_{A_i\neq A_{best}} \bm{1} \{Y(A_i,A_{Best}) > 0\} \leq (K-1)$, where $A_{best}$ represents the predicted best frame and $\bm{1}\{\cdot\}$ is the indicator function. We show the percentage of bursts among the whole test set that hit Top-K respectively.

\subsection{Implementation Detail}
Our system was implemented using the Caffe framework, trained on a NVIDIA Titan X GPU. We use standard SGD with the momentum 0.9 and weight decay 0.0005. Initial learning rate is set to 0.001, dropping by a 0.1 gamma every 12000 iterations, and a total of 100000 iterations. Each mini-batch contains 35 image pairs sampled randomly from different bursts. Aside from the layers in the head, all other layers from the backbone network are fine-tuned from ImageNet pre-trained weights. Following standard technique,  during the training we do random cropped sampling and augment the training set by simple mirroring. During testing, only center cropping is used. The training typically takes 6 hours to converge when either the learning rate drops below $1e^{-8}$ or the validation accuracy stay the same for a few epochs.

\begin{table}[!t]
\centering
\begin{tabular}{c|c|c|c|c|c|c|c|c}
\hline
\textbf{Backbone}&\textbf{Head} & \textbf{C'} & \textbf{\begin{tabular}[c]{@{}c@{}}Size (MB)\end{tabular}} & \textbf{Top 1} & \textbf{Top 2} & \textbf{Top 3}& \textbf{\begin{tabular}[c]{@{}c@{}}Pairwise\end{tabular}} & \textbf{GFlop} \\ \hline

ResNet-152 & A & 0    & 222.5              & 65.5        & 79.1       & 85.6 & 76.7 & 11.4       \\   \hline

GoogleNet & A & 0    & 39.3              & 65.0        & 79.5       & 86.1 & 76.0 & 1.6       \\   \hline
\multirow{10}{*}{SqNet} & A & 0 & 2.8 & 63.9 & 78.6 & 85.4 & 75.6 & \multirow{10}{*}{0.29} \\ \cline{2-8}
& B & 0 & 2.8 & 64.1 & 78.2 & 85.2 & 75.2 &	\\ \cline{2-8}
& C & 5 & 2.8 & 64.2 & 79.1 & 86.0 & 75.6 & \\ \cline{2-8}
& C+G & 5 & 2.8 & 65.6 & 80.3 & 86.8 & 77.2 & \\ \cline{2-8}
& C & 20 & 2.8 & 64.3 & 79.0 & 85.9 & 75.4 & \\ \cline{2-8}
& C+G & 20 & 2.8 & 65.9 & 80.3 & 87.0 & 77.2 & \\ \cline{2-8}
& C & 50 & 2.9 & 65.3 & 79.6 & 86.5 & 76.2 & \\ \cline{2-8}
& C+G & 50 & 2.9 & 65.9 & 80.3 & 86.9 & 77.1 & \\ \cline{2-8}
& C & 100 & 3.0 & 64.6 & 78.8 & 85.8 & 76.0 & \\ \cline{2-8}
& C+G & 100 & 3.0 & 65.7 & 80.0 & 86.9 & 77.4 & \\ \hline

\multirow{10}{*}{SqNet-4}& A & 0 &0.46 & 60.9 & 76.1 & 83.72 & 72.2 & \multirow{10}{*}{0.17} \\ \cline{2-8}
& B & 0 & 0.46 & 60.8 & 75.8 & 83.1 & 73.6 & \\ \cline{2-8}
& C & 5 & 0.47 & 62.3 & 77.3 & 85.1 & 74.5 & \\ \cline{2-8}
& C+G & 5 & 0.47 & 64.1 & 78.8 & 86.2 & 75.9 & \\ \cline{2-8}
& C & 20 & 0.48 & 62.8 & 77.4 & 84.9 & 74.3 & \\ \cline{2-8}
& C+G & 20 & 0.48 & 63.9 & 78.7 & 86.0 & 75.8 & \\ \cline{2-8}
& C & 50 & 0.51 & 63.4 & 78.4 & 85.7 & 74.8 & \\ \cline{2-8}
& C+G &50& 0.51 & 64.0 & 78.8 & 86.0 & 76.0 & \\ \cline{2-8}
& C & 100 & 0.56 & 62.5 & 77.1 & 84.7 & 74.1 & \\ \cline{2-8}
& C+G & 100 & 0.56 & 64.1 & 78.9 & 86.2 & 75.7 & \\ \hline
\multirow{10}{*}{SqNet-6} & A &0& 0.10 & 60.3 & 75.2 & 83.2 & 72.1 & \multirow{10}{*}{0.10}\\ \cline{2-8}
& B &0& 0.10 & 60.3 & 75.6 & 83.4 & 72.3 &\\ \cline{2-8}
& C & 5 & 0.10 & 60.3 & 75.3 & 83.1 & 72.1 & \\ \cline{2-8}
& C+G & 5 & 0.10 & 61.4 & 76.4 & 83.8 & 73.1 & \\ \cline{2-8}
& C & 20 & 0.11 & 60.9 & 75.9 & 83.7 & 72.3 & \\ \cline{2-8}
& C+G & 20 & 0.11 & 61.5 & 76.5 & 84.0 & 73.1 & \\ \cline{2-8}
& C &50& 0.12 & 61.0 & 75.9 & 83.6 & 73.2 & \\ \cline{2-8}
& C+G & 50 & 0.12 & 61.9 & 76.8 & 84.2 & 73.5 & \\ \cline{2-8}
& C & 100 & 0.15 & 61.1 & 75.9 & 83.5 & 72.4 & \\ \cline{2-8}
& C+G & 100 & 0.15 & 62.1 & 76.9 & 84.4 & 73.5 & \\ \hline

\end{tabular}
\caption{Results of the detailed ablation studies for the proposed head between \textbf{With} and \textbf{Without} adversarial regularization (G) when varying the number of relative attributes $C'$. All the metrics are measured as percentages ($\%$). As can be seen, adding G always improves the performance for all backbone and all $C'$, indicating the effectiveness of the adversarial regularization loss when training the ranking model}
\label{table:heads}
\end{table}

\subsection{Ablation Study}

\subsubsection{The effects of attribute-aware head design}

To study the efficiency and effectiveness of ``Head-C", we need to choose a foundation layer where it sits on. We tried three different versions of SqueezeNet \cite{squeezenet} by varying the number of trimmed ``Fire" layers. ``SqNet-4'' denotes a trimmed SqueezeNet with the top 4 fire layers removed. Like-wise, ``SqNet-6'' only keeps the layers from bottom to the third ``Fire'' layer block. We let ``SqNet" denote the full SqueezeNet that keeps the layers from bottom up to the last `'Fire" layer block. Clearly, because of the max-pooling layer, the top layer of these three backbone networks are different in terms of the shape. For example, ``SqNet-4'' outputs a feature map of size $[N$x$256$x$28$x$28]$. For each of the three SqueezeNet versions, we trained different models for each different head. In ``Head-C'', we also empirically set 4 different values for $C'$ and train separate models accordingly to study the effects of the attribute number in the intermediate layer. For this ablation\footnote{Unless otherwise noted, our ``Head-C'' is by default alway trained with ``G''} study only, we use C+G to represent ``Head-C'' was trained with the adversarial regularization loss, and use C to represent its counterpart without G. 

As expected, for any attribute number $C'$ and whichever backbone it sits on, compared with ``Head-A" and ``Head-B",``Head-C" only adds negligible extra FLOPs and model size but significantly boost the performance of accuracy under all different metrics, for both with and without GAN. This can be seen in Table \ref{table:heads}. For example, under ``SqNet-4'' and when $C'=5$, the Top-1 accuracy improves $3.2\%$ and $3.3\%$ from ``Head-A" and ``Head-B'' to ``Head-C" respectively, while only adding $2$K GFlops and $5$KB model size, which is negligible compared with the backbone. 

Even without the adversarial regularization loss(implemented by G), compared with Head-A and Head-B, adding the intermediate layer in Head-C seems to be always better for whichever $C'$, as shown in Table \ref{table:heads}. This validated our hypothesis that quickly reducing the features to one final score could loss much useful information, indicating adding the intermediate layer to preserve the relative attributes for ranking is effective. However, a larger $C'$ is not necessarily always better.   

To see how important the adversarial loss for the final performance in ``Head-C", we trained all the counterpart models without GAN. As can be seen in Table \ref{table:heads}, adding the adversarial loss during the training consistently improves the performance for all three back-bone nets and all the $C'$(100,50,20,5) we tried. For example, when $C'$ is fixed to 100, the gains that come from adding G are $1.1\%$,$1.6\%$ and $1\%$ for SqNet, SqNet-4 and SqNet-6 respectively. Interestingly, when $C'$ is reduced to 5, the gain by adding GAN is even more, the improvements are $1.4\%$, $1.8\%$ and $1.1\%$ respectively. All those studies indicate that GAN seems to be more effective for small backbone models and small number of latent attributes($C'$) in ``Head-C''.

Another interesting trend we can find is that the gain coming from the head optimization seems to be more economic compared with the gain coming by increasing the network capacity. For example, when $C'=50$, ``SqNet-6'' with ``Head-C'' ($61.9\%$) performs even better than ``SqNet-4'' with ``Head-B''($60.8\%$). However, the later model is 4 times larger and 1.7 times more computationally costly in terms of FLOPS. Similarly, ``SqNet-4'' with ``Head-C'' also performs slightly better than the full``SqNet" with ``Head-A'' ($64.1\%$ VS $63.9\%$), but again the later model is 4 times larger and 1.7 times more computationally costly. This may indicate that attribute-aware head design indeed encodes more intrinsic relative attributes that make the ranking function easier to learn.

\begin{figure*}[!t]
\centering
\includegraphics[width=0.98\linewidth]{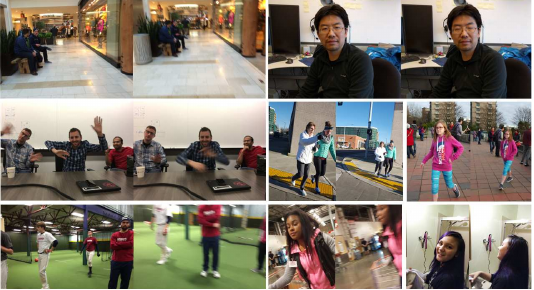}
\caption{A few testing examples that hit top-1 accuracy. The left(right) column represent the best(worst) frames for each burst respectively.}
\label{fig:gallery}
\end{figure*}

\begin{figure*}[h!]
\centering
\includegraphics[width=0.98\linewidth]{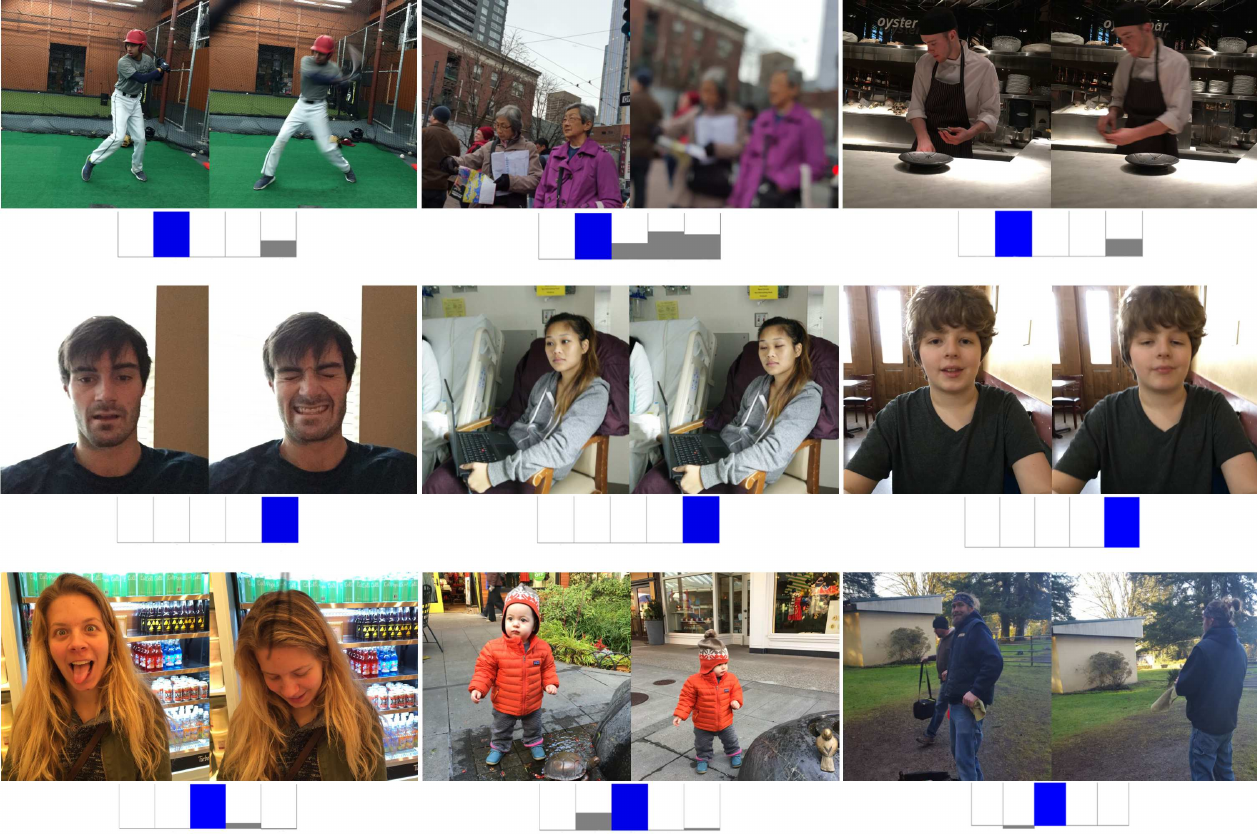}
\caption{Three different clusters representing relative attributes for sharpness, eye opened, and human pose respectively. Under each image pair (left is the best, right is the worse), we visualize the attribute difference gap histogram.}
\label{fig:cluster}
\end{figure*}

\subsubsection{The effects of back-bone}
For big networks such as GoogleNet\cite{GoogleNet} and ResNet\cite{resnet}, it seems to be infeasible to make it real-time on mobile devices not to say its huge model size(i.e.,$~222$M for ResNet-152). Hence, our main focus is to study the performance of our design on small backbone network such as SqueezeNet. Nevertheless, we still train both GoogleNet and ResNet as the backbones for comparisons. As shown in Table \ref{table:heads}, when using the same ``Head-A", GoogleNet and ResNet indeed hit higher accuracy compared with SqNet-4 and SqNet-6, the gain however is very minor given the capacity difference. However, a SqNet with ``Head-C" can even achieve a higher accuracy than ResNet-152 with ``Head-A", even though SqNet is only 50x smaller. We have noticed that, a deeper network does not always generate better performance of accuracy for our photo ranking problem. We argue that, unlike image classification task, relative attribute learning for ranking may not require very fine-grained features to distinguish between a cat versus a dog. So a very large capacity network, such as ResNet-152 may have high risk of over-fitting.  However, when the capacity of backbone is less than a threshold, for example 3M for SqNet, the depth of backbone starts to be relatively important, for example, SqNet-6 is almost $4\%$ worse than SqNet. We argue that, 4M could be a reasonable model size for an camera application on mobile devices, therefore, it is more critical to design a novel head to improve the performance for the small backbone models, which is exactly our focus in this work.

\subsubsection{What features we learned?} 
Figure \ref{fig:teaser} shows one typical burst as well as the ranking result predicted by our model. Figure \ref{fig:gallery} shows a gallery of testing results where our predictions all hit the Top-1 accuracy. For the sake of comparison, we only show the best and worse frame within each burst. Clearly, our model favors more opened eyes vs. closed eyes, more saliency subject (like the girl), better body pose (i.e., kid), and sharpness. We chose SqNet-4, with $C'=5$ in ``Head-C'' as the final model, and visualize the attributes difference gap histogram for image pairs of test set. In Figure \ref{fig:cluster}, we show a few such examples, from where we can find a clear clustering effect, i.e, the fifth dimension in the attribute space looks like focuses more on the attribute of eye openness, while the second and third dimension focus more on sharpness and human pose respectively. 

\subsection{Comparison with Prior Work}

\begin{table}[t]
\centering
\begin{tabular}{c|c|c|c|c}
\hline
\multicolumn{2}{c|}{\textbf{Model}}                                & \textbf{\begin{tabular}[c]{@{}c@{}}Size\\ (MB)\end{tabular}}    & \textbf{Pairwise} & \textbf{Flop} \\ \hline
\multirow{1}{*}{Baseline \cite{Chang:2016:ATF}}                     & VGG-16          & 514.2  & 73.2    & 15.5      \\ \hline
\multicolumn{1}{c|}{\multirow{2}{*}{Ours}} & VGG-16$^*$+C          & 47.2        & 73.1 & 15.5             \\ \cline{2-5} 
\multicolumn{1}{c|}{}                      & SqNet-4+C & 0.51  & 72.9    & 0.17     \\ \hline
\end{tabular}
\caption{Comparison with \cite{Chang:2016:ATF} on their dataset. VGG-16$^*$ represents a trimmed VGG with all fully connected layers removed. Note that, our model is about 90x smaller than the trimmed VGG net and 1000x smaller than the baseline model used in \cite{Chang:2016:ATF}.} 
\label{tab:triagle}
\end{table}
In photo triage work \cite{Chang:2016:ATF}, they have shown that CNN-based relative learning is very efficient, and can beat all the competitors including the ones that rely on hand-crafted features. So we only chose to compare with their CNN approach.

Although they target a different yet related problem and they do not support burst session data, technically, the approach can still be used for burst photo ranking problem. Their head design can be seen as the ``baseline" head in Figure \ref{fig:head_design}. Although, the backbone is shared, the ranking always needs to take a feature vector pair as input to get the result, which may not be very efficient for real-time capture, as we cannot run all the frames within a burst in parallel. The time complexity of getting the full rank is $O(N^2)$. Whereas in our design, during the runtime, each image can be run independently to get the final score. Following their design principle, we conducted side-by-side comparison by training another model that sits on ``SqNet-4'' and using their header that shows in Figure \ref{fig:head_design}. On our dataset, we get $62.0\%$ Top-1 accuracy, which is $2.1\%$  worse compared with our design ``Head-C'' as shown in Table \ref{table:heads}. 

We further trained another two models with our ``Head-C" that sits on a trimmed VGG (with FC layers removed) and ``SqNet-4'' respectively on their Triage dataset. Compared with their VGG-16 baseline, our proposed model is up to 1000x smaller in model size and 90x faster, with relatively the same accuracy, as shown in Table \ref{tab:triagle}. This again indicates that our design is both effective and efficient, and is general for photo ranking beyond burst data.

\subsection{Comparison with Native Cameras}
\begin{figure}[!t]
\centering
\includegraphics[width=0.5\linewidth]{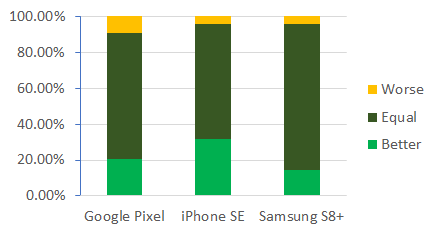}
\caption{Comparison with existing best of burst algorithms}
\label{fig:pairwise_label_dist}
\end{figure}

We further conducted an informal user study though comparing our model with the built-in best of burst algorithm in Sam-sung Galaxy S8 Plus, Google Pixel and iPhone SE respectively
We used the burst capture mode in the system camera app to capture around 200 bursts of ten to thirty frames. 
Post processing on these devices' native camera application picks the best frame from the burst automatically. We ran the same bursts through our technique and picked the top scoring frame. To quantify the quality of the results of our technique, we conducted a five person blind A/B test to find if a user likes the system default best frame or the best frame from our technique. We averaged the responses per image-pair and rounded to the nearest option (better, equal or worse). As shown in Figure \ref{fig:pairwise_label_dist}, we see that results from our technique are clearly preferred by users when compared to the native best of burst algorithm, on both Android and iOS. We noticed that the native algorithms mostly take into account image blur without accounting for facial expressions and pose. We also deployed our system into both iPhone 7 and Google Pixel phones without aggressively engineering low level optimization. For the model ``SqNet-4" and ``Head-C", the runtime only takes 13ms on iPhone 7 and about 26ms on Google Pixel phone.

\section{Conclusion}
In this work, we have presented a real-time burst moment capture system based on deep learning. We formulate the problem as a relative learning problem for ranking. Currently, we consolidate the annotation of human label by simple averaging. Thus we only expect the model to learn general preferences. As a future work, one may consider to apply advanced techniques to learn personalized models for photo ranking.

\bibliographystyle{splncs}
\bibliography{egbib}
\end{document}